\documentclass[letterpaper, 10 pt, conference]{ieeeconf}

\IEEEoverridecommandlockouts                              

\usepackage{times}
\usepackage{soul}
\usepackage{url}
\usepackage[colorlinks]{hyperref}
\usepackage[utf8]{inputenc}
\usepackage{graphicx}
\usepackage{amsmath}
\usepackage{booktabs}
\usepackage{algorithm}
\usepackage{algorithmic}
\usepackage{color}

\makeatletter
\IEEEtriggercmd{\reset@font\normalfont\scriptsize}
\makeatother
\IEEEtriggeratref{1}

\renewcommand{\mathbf}{\boldsymbol}

\newcommand{\N}{\mathcal{N}}

\newcommand{\nop}[1]{}

\def\eg{\emph{e.g.}}
\def\ie{\emph{i.e.}}

\title{\LARGE \bf
Towards Robust Human Trajectory Prediction in Raw Videos
}

\author{Rui Yu and Zihan Zhou*
\thanks{*R. Yu and Z. Zhou are with College of Information Sciences and Technology, Pennsylvania State University, University Park, PA 16802, USA {\tt\small \{rzy54, zuz22\}@psu.edu}}%
\thanks{This work is supported by NIH Award R01LM013330.}
}

\begin{document}

\maketitle
\thispagestyle{empty}
\pagestyle{empty}


\begin{abstract}

Human trajectory prediction has received increased attention lately due to its importance in applications such as autonomous vehicles and indoor robots. However, most existing methods make predictions based on human-labeled trajectories and ignore the errors and noises in detection and tracking. In this paper, we study the problem of human trajectory forecasting in raw videos, and show that the prediction accuracy can be severely affected by various types of tracking errors. Accordingly, we propose a simple yet effective strategy to correct the tracking failures by enforcing prediction consistency over time. The proposed ``re-tracking’’ algorithm can be applied to any existing tracking and prediction pipelines. Experiments on public benchmark datasets demonstrate that the proposed method can improve both tracking and prediction performance in challenging real-world scenarios. The code and data are available at \url{https://git.io/retracking-prediction}.

\end{abstract}

\section{Introduction}

Driven by emerging applications such as autonomous vehicles, service robots, and advanced surveillance systems, human motion prediction has received increased attention in recent years~\cite{rudenko2020human}. In the literature, most studies apply regression models on the subjects' past trajectories to recursively compute the target positions several time steps into the future. Some traditional methods are based on motion models such as linear models, Kalman filters~\cite{KimGLWLLM15}, Gaussian process regression models~\cite{EllisS009}, and social force models~\cite{Helbing95}. Recently, data-driven deep learning models such as Long Short-Term Memory (LSTM) have been shown to achieve higher prediction accuracies, thanks to their ability to modeling complex temporal dependencies and human interactions in the sequential learning problem~\cite{AlahiGRRLS16,GuptaJFSA18}. 

In the aforementioned methods, the subjects' past movements, which serve as the input, are assumed to be given. 
However, in real-world scenarios, the system often needs to first estimate the past trajectories from raw video data. 
For example, for an autonomous vehicle to safely and efficiently navigate in city traffics, it is necessary to understand and predict the movement of pedestrians from the video stream captured by its on-board cameras. Since most prediction methods do not explicitly consider the errors and uncertainties incurred by detection and tracking, directly applying them often leads to \emph{inconsistent predictions} over time. 

\begin{figure}[t]
\centering
\begin{tabular}{cc}
\hspace{-3mm}\includegraphics[height=0.72in]{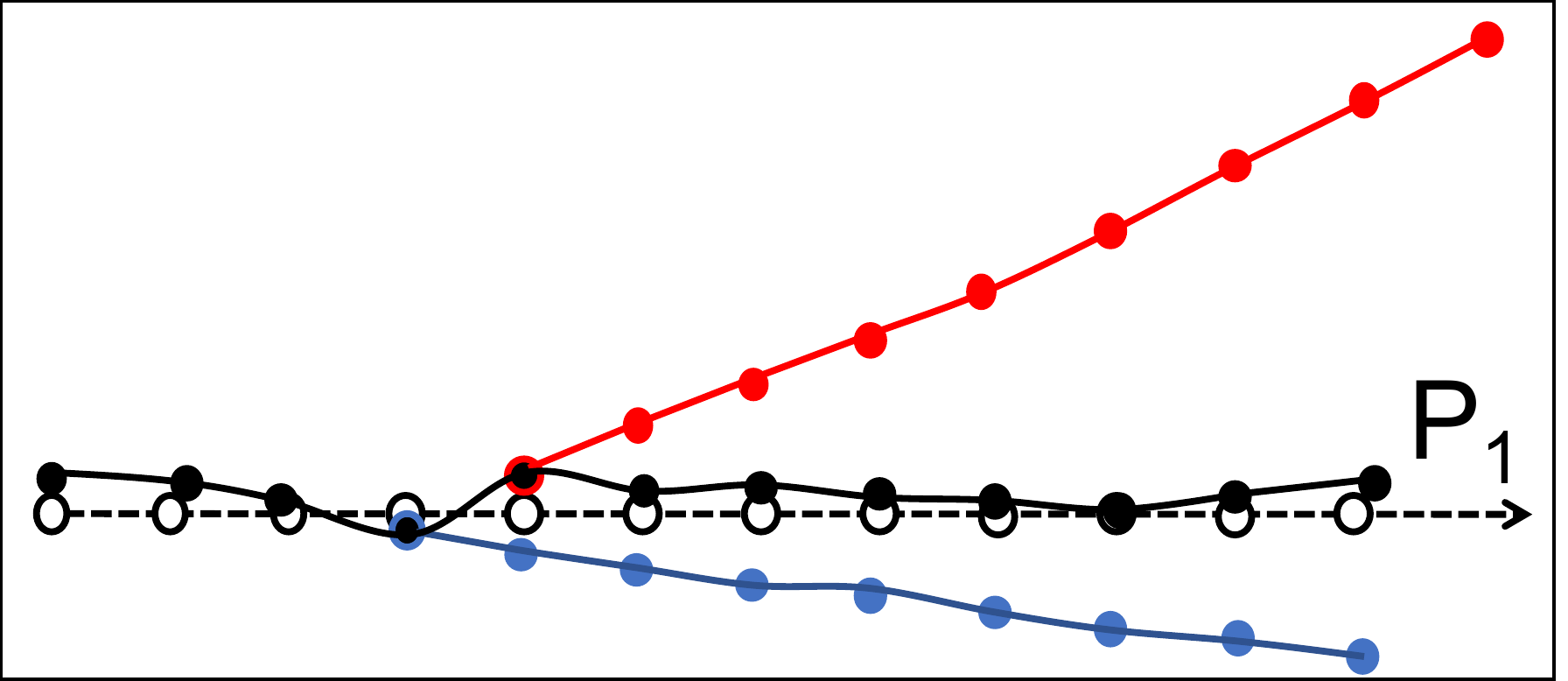} &
\hspace{-3mm}\includegraphics[height=0.72in]{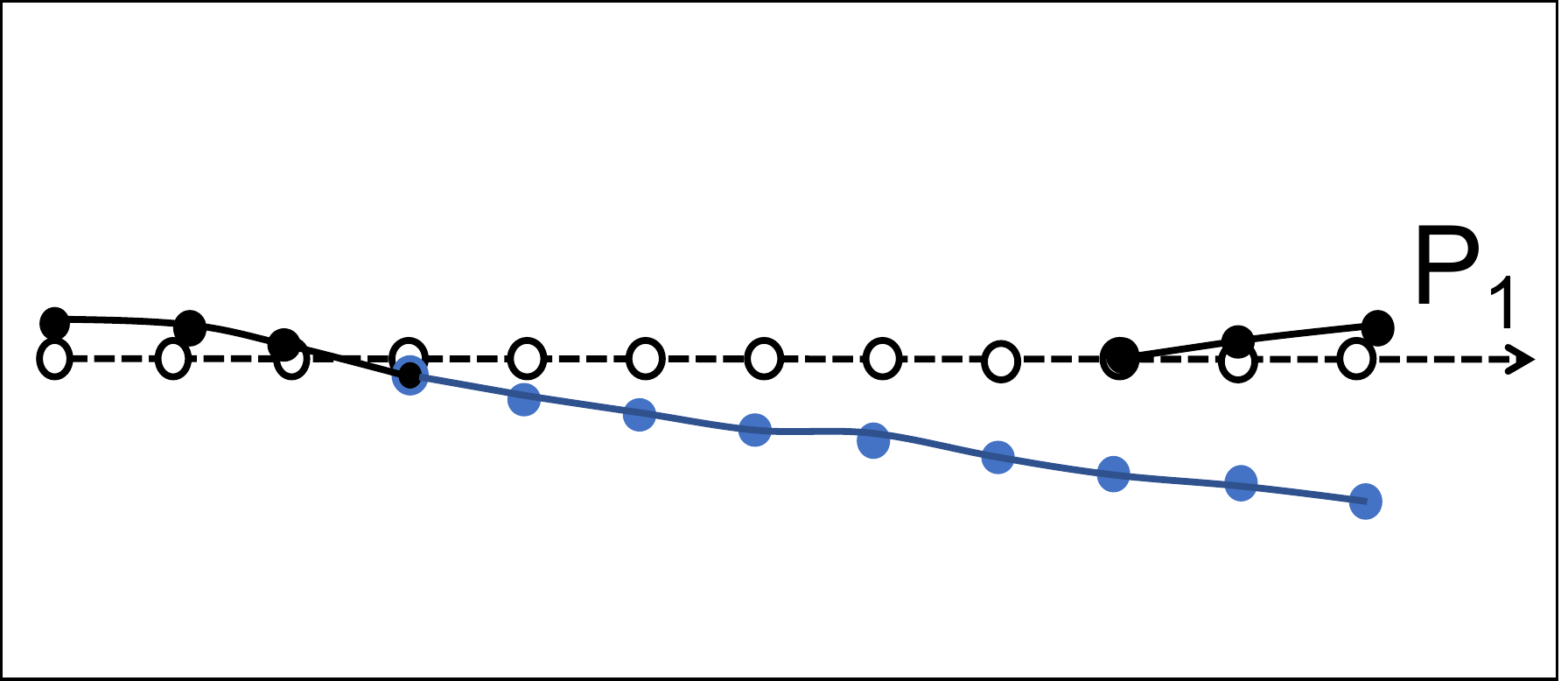} \\
{\small(a) Noisy track} & {\small(b) Missed targets}
\end{tabular}
\begin{tabular}{cc}
\hspace{-3mm}\includegraphics[height=0.715in]{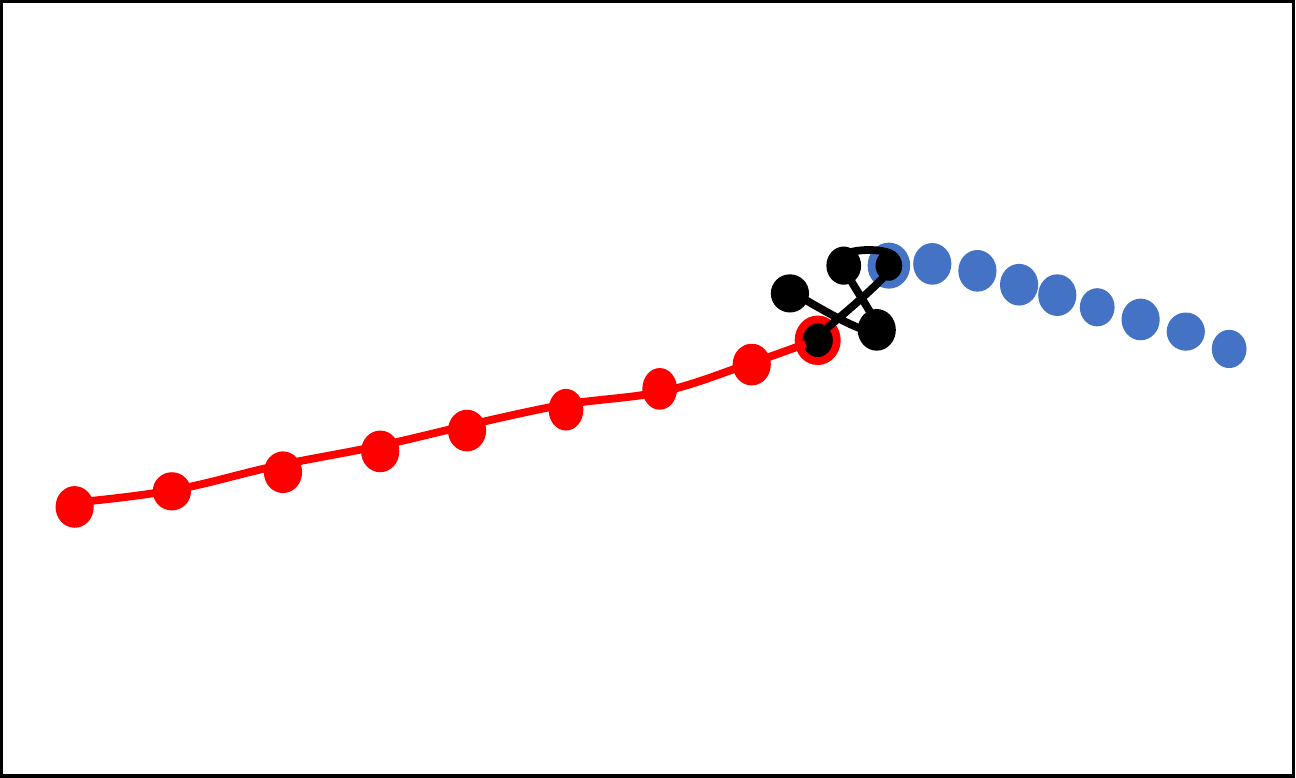} &
\hspace{-3mm}\includegraphics[height=0.715in]{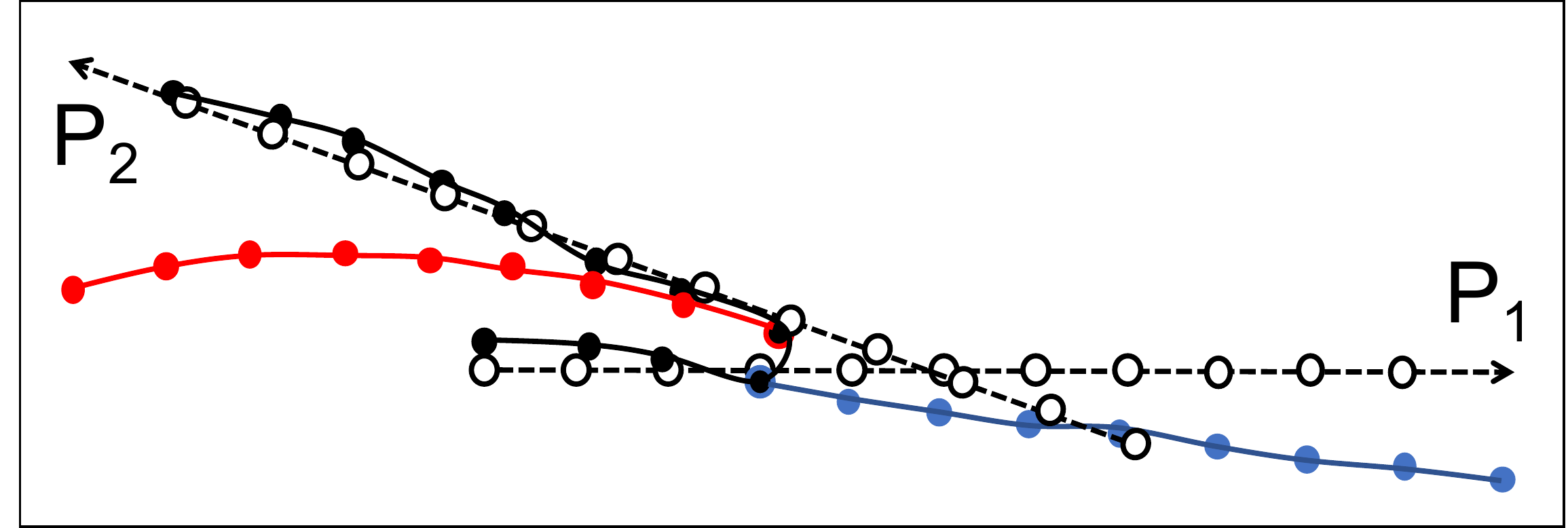} \\
{\small(c) Spurious track} & {\small(d) ID switch}
\end{tabular}
\vspace{-2ex}
\caption{Common causes of inconsistent human trajectory predictions in raw videos. In each figure, we show the ground truth trajectory (black circles), tracker outputs (black dots), predictions at time $t$ (blue dots), and predictions at time $t+1$ (red dots).}
\label{fig:problem}
\end{figure}

In Fig.~\ref{fig:problem}, we illustrate several common cases of inconsistent predictions in raw videos. As seen in Fig.~\ref{fig:problem}(a), the estimated tracks may not strictly follow the ground truth (GT) trajectory of a subject. Because of the apparent change of direction due to the noisy estimation, the predictions at time $t$ could be very different from those at time $t+1$. Besides, the tracking results may contain various types of errors including missed targets, spurious tracks, and ID switches. As a result, the predictions at consecutive time steps (if exist) could differ significantly (Fig.~\ref{fig:problem}(b)-(d)).


\setlength{\textfloatsep}{4pt}

In this work, we study the problem of human motion prediction in raw videos. We show empirically that, due to the aforementioned reasons, there is a significant performance gap between prediction in raw videos and that using manually labeled human trajectories, especially when reliable detection and tracking is difficult to obtain (\eg, small objects, crowded scenes, camera movements).


As an attempt to bridge this gap, we propose a simple yet effective strategy to improve the prediction performance in raw videos by enforcing \emph{temporal prediction consistency}, a property largely ignored by prior work. Specifically, given the results obtained by any tracking algorithms, we first apply a smoothing filter to the estimated tracks. Then, we repeatedly run the prediction model at every tracked location, reconstruct a new track for each human subject by comparing the similarity of prediction results for points at consecutive time steps, and generate the final predictions using the new track. The advantage of our ``re-tracking'' algorithm is three-fold. First, compared to the original tracks, the results obtained by our algorithm have significantly fewer missed targets, spurious tracks, and ID switches. Second, the predicted future trajectories are less sensitive to the noises in the original tracks, thus are more accurate. Third, our ``re-tracking'' algorithm is independent of the tracking and prediction methods, thus can be applied to any existing tracking-prediction pipeline as a standalone module.

We conduct experiments on popular human motion prediction benchmarks. Since our goal is to predict the movement of \emph{all} pedestrians in the video frame, we systematically evaluate the proposed method against baselines in terms of both tracking and prediction. We show that our method can simultaneously improve the tracking and prediction accuracies by addressing the issues shown in Fig.~\ref{fig:problem}. For example, it can reduce the number of ID switches by more than $65\%$ on the SDD test sets~\cite{robicquet2016learning}.

In summary, the contributions of this work are as follows. (1) We study the problem of human trajectory prediction in raw videos, which has received less attention in the literature so far. We analyze the relationship between tracking and prediction, and illustrate the challenges in achieving temporally consistent predictions for this problem.
(2) We propose a ``re-tracking’’ algorithm with the goal of improving temporal prediction consistency. We show that it leads to better tracking and prediction performance on public benchmark datasets. 

\section{Related Work}



\subsection{Human Trajectory Prediction}

There is rich literature on understanding and predicting human motion from visual data. We refer readers to~\cite{rudenko2020human} for a comprehensive review of existing methods. Below we provide a brief overview of recent data-driven methods which are most relevant to our work.


Inspired by the recurrent neural network (RNN) models for sequence generation~\cite{Graves13}, \cite{AlahiGRRLS16} first proposed to use the RNN to solve the human trajectory prediction problem. 
Following their work, various deep networks were developed by integrating techniques such as attention models~\cite{VemulaMO18,sadeghian2019sophie}, generative adversarial networks (GAN)~\cite{GuptaJFSA18}, pose estimator~\cite{HasanSTBGC18}, variational autoencoder (VAE)~\cite{MangalamGALAMG20}, graph neural networks (GNN)~\cite{KosarajuSM0RS19,MohamedQEC20}, and Transformer networks~\cite{YuMRZY20}. These methods represent human subjects as 2D points on the ground plane and only employ static environmental images, thus ignore the temporal context in the raw videos.


Several works use raw video frames for human motion prediction. \cite{liang2019peeking} proposed a multi-task network to jointly predict future paths and activities of the pedestrians from raw videos. But the ground truth bounding boxes for the past time steps are still given. \cite{YagiMYS18} proposed to predict pedestrian locations in first-person videos. In their new dataset, some heuristics are used to choose successfully detected human locations and poses as ground truth. \cite{BhattacharyyaFS18} proposed a two-stream framework with RNN and CNN to jointly forecast the ego-motion of vehicles and pedestrians with uncertainty. In both works, detection and tracking errors were treated as a minor nuisance to demonstrate the robustness of the proposed method. Recently, however, it is argued in~\cite{MaZCYLM20} that extracting human trajectories from raw videos remains a challenging problem. It further leveraged unlabeled videos for training deep prediction models. However, their evaluation is still based on ground truth observations. 

\subsection{Joint Tracking and Prediction}
\label{sec:related-joint}

Human motion models have also been used to assist tracking. A linear model with constant velocity assumption~\cite{BreitensteinRLKG09} is by far the most popular model. However, real-world human movement patterns are often complicated. In~\cite{YangN12}, a non-linear model was used to handle the situation that the targets may move freely. To further consider the interaction among targets, \cite{PellegriniESG09,YamaguchiBOB11} proposed to use the social force model~\cite{Helbing95}. The most closely related work to ours is~\cite{FernandoDSF18}, which proposed a ``tracking-by-prediction'' paradigm. 
It compares short-term predictions (\ie, next two frames) with each detection for data association. In our work, we directly compare long-term predictions using the trajectory-wise Mahalanobis distance for data association, with a focus on generating temporally consistent predictions.

Another recent line of work~\cite{LuoYU18,CasasLU18,GLU20,LiangYZCH0U20,CasasGSLLU20,LiYLZRSU20,abs-2010-00731} studies joint 3D detection, tracking, and motion forecasting of vehicles in traffic scenes from 3D LiDAR scans. These methods are similar to ours as they do not rely on past ground truth trajectories for motion forecasting. For prediction evaluation, they use a fixed IoU threshold (\ie, 0.5) to associate detections with ground truth bounding boxes in the current frame, and report the Average Displacement Error (ADE) and Final Displacement Error (FDE) at a fixed recall rate (\eg, 80\% as in~\cite{LiYLZRSU20}).

The influence of detection and tracking on motion forecasting is investigated more carefully in~\cite{weng2020inverting}. Instead of associating estimated tracks to past GT trajectories, it directly matches predicted trajectories with future GT trajectories, and reports the ADE-over-recall and FDE-over-recall curves. But such measures are incomplete in the sense that it does not consider all types of errors in the pipeline. For example, it is possible to simultaneously achieve high recall and low ADE by generating a large number of hypotheses, but also introducing many false positives (\ie, ghost trajectories).

\section{Problem Analysis and Method Overview}

As mentioned before, given a raw video, we wish to predict the future movement of all human subjects in the scene. Formally, at any time step $t$, a person in the scene is represented by his/her xy-coordinates $(x_t, y_t)$ on the ground plane. Following previous work, we formulate the task as a sequence generation problem. Given any time step of interest $t$, let $\Omega_t$ be the set of all human subjects in the video frame $I_t$. Our goal is to predict their positions for time steps $ t+[1:t_{pred}]$, $\forall s\in \Omega_t$. 

In this work, we consider a general two-stage approach to tackle this problem. The first step is to perform \emph{detection and tracking} to obtain a set of object instances $\Omega_{t}'$ in the frame $I_t$. Each instance $s'\in \Omega_{t}'$ is associated with a sequence of coordinates representing the subject's past movement $\{(x'_{t-t_{obs}},y'_{t-t_{obs}}), \ldots, (x'_t,y'_t)\}$. Next, for each instance, a prediction model (\eg, LSTM) takes the movement history as input and generates its future movement predictions. In practice, however, the set of tracked instances $\Omega_{t}'$ suffers from issues including missed targets, spurious tracks, and ID switches. Further, the past trajectories may contain noises and errors. As a result, we often observe inconsistency in the predicted trajectories at consecutive time steps (Fig.~\ref{fig:problem}). 

\begin{figure}[t]
\centering
\includegraphics[height=0.55in]{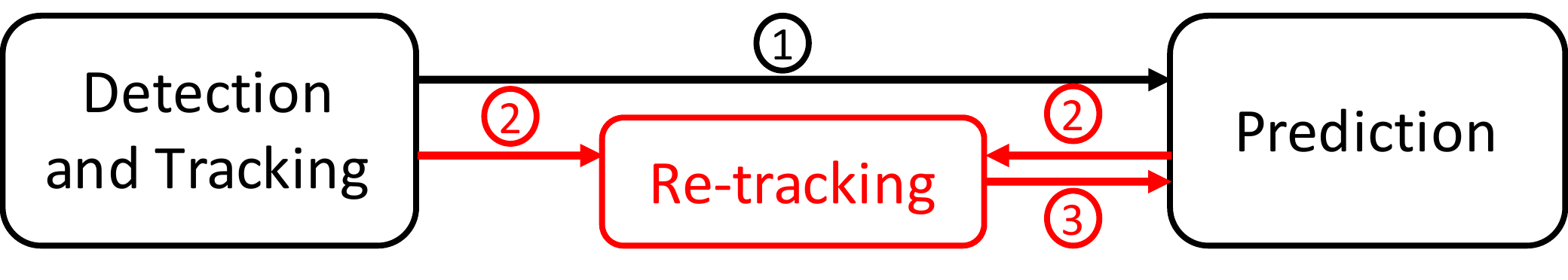} 
\vspace{-2ex}
\caption{Proposed pipeline. (1) Given the outputs of an existing tracking method, we predict a subject's future movement at each tracked location. (2) The re-tracking module uses the predictions to build a new set of tracks. (3) With the new tracks, we are able to generate more consistent predictions in raw videos.}
\label{fig:pipeline}
\end{figure}

In view of the inconsistent predictions, we ask the following question: \emph{Is it possible to construct a new set of tracks with which the prediction model will generate more consistent results?} In this work, we show that it is possible, by considering future predictions in tracking. The resulting algorithm is a standalone method that uses a prediction model to improve the estimated tracks (Fig.~\ref{fig:pipeline}). For any time $t$, it constructs a new set of object instances $\Omega_{t}''$, where each $s''\in \Omega_{t}''$ is associated with a new sequence of locations. We call it a ``re-tracking'' method because it operates on, and further refines, the outputs of existing tracking methods. In the next section, we describe our method in detail.

\section{Proposed Re-tracking Method}

As input to our method, we assume that we are given a set of trajectories $\{S_1, \ldots, S_N\}$, where $S_k = \{(x_t^k, y_t^k)\}_{t = t_a^k}^{t_b^k}$. Here, $t_a^k, t_b^k$ denote the starting and ending time steps, respectively. We disregard the identity of the trajectory and treat each location $(x_t^k, y_t^k)$ as an independent observation at time $t$. Let $O_t$ be the set of all observations in time $t$. Our goal is to associate the observations across different time steps to recover the subjects' trajectories.

To build the new tracks, we explicitly take the prediction performance into account. First, we filter the original tracks to improve the prediction consistency. Second, we compare the predictions made at different time steps and use the differences as a cue to recover missed targets and remove outliers in the tracks. The design of our re-tracking method is based on the following simple ideas:

\smallskip
\noindent{\bf Smoothing the input sequence:} We observe that, at any time $t$, the most recent relative motion (or instantaneous velocity) $\Delta_t^k = (x_t^k-x_{t-1}^k, y_t^k-y_{t-1}^k)$ is a dominant predictor for a subject's future movement. In~\cite{SchollerALK20}, a similar observation has also been made, regardless of the prediction model. In practice, this suggests that small perturbations to the subject's estimated location could have a significant impact on the prediction outputs (Fig.~\ref{fig:problem}(a)). Based on this observation, we propose to use Holt–Winters method~\cite{holt60}, a classic technique in time series, to smooth the instantaneous velocities in an online manner. The smoothed sequences are then used to predict the subject's future movement.

\smallskip
\noindent{\bf Repeated predictions:} In most prior work on human motion prediction, a subject's trajectory is partitioned into small segments on which the prediction is performed given the ground truth locations for the first $t_{obs}$ time steps. In practice, however, the past $t_{obs}$ locations may not always be available. In this work, we propose to make a prediction from \emph{every} observation (\ie, tracked location) whenever it is possible. Obviously, this will lead to a lot of redundant predictions, but also comes with two benefits:  First, it enables prediction of the subject's movement even if detection and tracking fail for some of the time steps, thus improves the prediction recall. Second, using the repeated predictions, we are able to build a new track for each subject that has fewer missed targets and outliers, as we explain next.

\smallskip
\noindent{\bf Re-tracking by prediction:} Since now each observation $o \in O_t$ is associated with a prediction $P_o = \{p_{t+1}^o, \ldots, p_{t+t_{pred}}^o\}$, we can group the observations based on the difference in predictions to re-build the trajectory of each human subject. Unlike most tracking methods which perform data association based on the bounding box distance in the \emph{current frame}, our method uses (long-term) future predictions. This allows us to connect observations across multiple time steps, and remove observations whose predictions are very different from the others (\ie, outliers).

\begin{algorithm}[t] \caption{Re-tracking by Prediction}\label{algo1}
\begin{algorithmic}[1]
\STATE \textbf{Input}: A set of observations $\{O_t\}_{t=1}^T$; max age $t_{max}$; matching distance threshold $d_{min}$;
\STATE \textbf{Initialize}: $M_a = \emptyset$; 
\FOR{each frame $t$}
    \STATE Perform Hungarian matching: $M_{m}, M_{um}, O_{um} = Hungarian(M_a, O_t)$;
    \FOR{each matched track $m\in M_{m}$}
        \STATE Smooth the associated observation $o$ as in Eq.~\eqref{eq:holt};
        \STATE Update $P_m$ with the smoothed observation;
        \STATE $a_m\leftarrow 0$;
    \ENDFOR
    \FOR{each unmatched track $m\in M_{um}$}
        \STATE $a_m\leftarrow a_m+1$;
    \ENDFOR
    \FOR{each unmatched observation $o\in O_{um}$}
        \STATE Start a new track $m = (P_o, 0)$ and add to $M_a$;
    \ENDFOR
    \FOR{each track $m\in M_{a}$}
        \IF{$a_m > t_{max}$}
            \STATE Remove $m$ from $M_a$;
        \ENDIF
    \ENDFOR
    \STATE \textbf{Output}: $M_a$;
\ENDFOR
\end{algorithmic}
\end{algorithm}

\subsection{Algorithm}

Now we describe our re-tracking algorithm in detail. The overall procedure is summarized in Algorithm~\ref{algo1}. In the algorithm, we maintain a set of active tracks $M_a$ at each time step. Each track $m\in M_a$ is associated with a prediction $P_m$ and an age $a_m$ and denoted by $m = (P_m, a_m)$.

\smallskip
\noindent{\bf Distance measure.} Given a track $m$ and an observation $o$, we need to compute the distance between them. To this end, We assume each predicted location in $P_o$ follows a Gaussian distribution: $p_u^o \sim \N\left(\mu_u^o, \Sigma_u^o\right), \forall u\in t+[1:t_{pred}]$. Similarly, for each prediction in $P_m$ we have $p_u^m \sim \N\left(\mu_u^m, \Sigma_u^m\right)$. The Mahalanobis distance between two distributions is:
\begin{equation}
    d(p_u^m, p_u^o) = \sqrt{(\mu_u^m - \mu_u^o)^T(\Sigma_u^m + \Sigma_u^o)^{-1}(\mu_u^m - \mu_u^o)}.
\end{equation}
Note that in the above Gaussian distribution, $\mu_u^o$ is simply the location predicted by the model, and $\Sigma_u^o$ is a diagonal matrix whose entries along the diagonal are equal to $(\sigma_u^o)^2 = (u-t) \times \sigma^2$. Here, $\sigma^2$ is a constant, and the coefficient $u-t$ represents our belief that the prediction becomes more and more uncertain into the future. The distribution for $p_u^m$ is defined in a similar way.
Then, we define the distance between $P_m$ and $P_o$ as:
\begin{equation}
\label{eq:traj_dist}
    d(P_m, P_o) = \frac{1}{|T_{m,o}|}\sum_{u\in T_{m,o}} d(p_u^m, p_u^o),
\end{equation}
where $T_{m,o}$ is the set of overlapping timestamps between the predictions $P_m$ and $P_o$ with $|T_{m,o}|\leq t_{pred}-1$.
Based on the distance measure in Eq.~\eqref{eq:traj_dist}, we use Hungarian algorithm~\cite{kuhn1955hungarian} for data association between $M_a$ and $O_t$.

\smallskip
\noindent{\bf Updating a track.} 
When a track $m$ is associated with a new observation $o$, we use the observation to update the track $m$ and generate a new prediction $P_m$. Recall that, because of the noises in the tracking results, directly using the observations as input to the prediction model may produce inconsistent predictions (Fig.~\ref{fig:problem}(a)). Therefore, we apply a smoothing filter to the estimated track. Note that, instead of directly smoothing the observed locations, we smooth the relative motion. This is because the most recent relative motion is shown to be a dominant predictor for future movement~\cite{SchollerALK20}. 

Specifically, let $\{o_{t_a}, o_{t_a+1}, \ldots, o_t\}$ be the sequence of observations associated with $m$ up to time step $t$. We first compute the relative motion $\Delta_u = o_u - o_{u-1}, \forall u\in [t_a+1 : t]$. Then, we use the Holt–Winters method (also known as double exponential smoothing) to recursively compute the smoothed motion:
\begin{align}
\label{eq:holt}
    \Delta'_t & = \alpha \Delta_t + (1-\alpha) (\Delta'_{t-1} + b_{t-1})  \nonumber\\
    b_t & = \beta (\Delta'_t - \Delta'_{t-1}) + (1-\beta) b_{t-1}
\end{align}
Note that a track $m$ may not have an associated observation at every time step. In such cases, we use a simple linear interpolation to recover the full time series. Finally, we use $\Delta'_t$ to reconstruct the past trajectory of the subject, which is then used as input to the prediction model to generate $P_m$. 


\section{Experiments}

\subsection{Experimental Settings}

\noindent{\bf Datasets.}
We evaluate the proposed method primarily on the \emph{Stanford Drone Dataset} (SDD)~\cite{robicquet2016learning}. SDD is a widely used benchmark for human trajectory prediction, containing traffic videos captured in bird's-eye view with drones. 
Following~\cite{sadeghian2019sophie,LiMT19}, we use the standard data split
~\cite{sadeghian2018trajnet} with 31 videos for training and 17 videos for testing.
During testing, we conduct pedestrian detection and tracking at 30 fps 
on all testing videos ($129,432$ frames in total) except for an extremely unstable one (\emph{Nexus-5} with $1,062$ frames), then evaluate the prediction performance at 2.5 fps to predict 12 future time steps (4.8 sec) as in previous works~\cite{sadeghian2019sophie,sadeghian2018trajnet}.

\smallskip
\noindent{\bf Evaluation metrics.} To systematically evaluate the performance of the pipeline, we ask the following two questions:
\begin{itemize}
    \item How many subjects are correctly tracked at any time $t$?
    \item For those tracked subjects, what are the differences between the predicted trajectories and ground truth? 
\end{itemize}

For the first question, we employ MOT metrics~\cite{Leal-TaixeMRRS15} of Identity F1 score ($IDF1$) and MOT Accuracy ($MOTA$):
\begin{equation}
    MOTA = 1 - \frac{\sum_t (FP_t + FN_t + IDSW_t)}{\sum_t GT_t},
\end{equation}
where $FP_t$, $FN_t$, $IDSW_t$, and $GT_t$ represent the number of false positives, false negatives, identity switches, and ground truth annotations at frame $t$, respectively.

For the second question, we use ADE to evaluate the performance of a prediction method, which is the average mean square error between the ground truth future trajectory and the predicted trajectory. In our problem, however, the inputs are the estimated tracks that we need to match the set of tracked instances $\Omega'_t$ with the set of all human subjects $\Omega_t$ in the video frame. For a pair of object instances $(s, s'), s\in \Omega_t, s'\in \Omega'_t$, we compute the distance of the pair as:
\begin{equation}
   d_{obs}(s,s') = \frac{1}{t_{obs}}\sum_{u=t-t_{obs}+1}^{t} (x_u-x'_u)^2 + (y_u - y'_u)^2.
\label{eq:association}
\end{equation}
Then, the pairwise distances are fed to the Hungarian algorithm to obtain a one-to-one correspondence between all the tracks and the ground truth. We consider a ground truth subject $s$ correctly matched with $s'$ if their distance is below a threshold $\tau$. Thus, we only compute the ADE on the set of correctly matched subjects $M$ as obtained in the tracking evaluation:
$ADE = \frac{1}{|M|} \sum_{(s,s')\in M} d_{pred}(s,s')$, where
\begin{equation}
    d_{pred}(s,s') = \frac{1}{t_{pred}}\sum_{u=t+1}^{t+t_{pred}} (x_u-x'_u)^2 + (y_u - y'_u)^2.
\end{equation}

Obviously, the choice of threshold $\tau$ has an impact on the prediction evaluation, because the prediction accuracy depends not only on the prediction model, but also on how much the input track deviates from the true location of the subject. 
By varying the value of $\tau$, we can obtain different association recalls and the corresponding ADE values, and then plot an ADE-over-recall curve.

\smallskip
\noindent{\bf Baseline and implement details.}
It is non-trivial to detect small objects from bird's-eye view~\cite{MaZCYLM20} that the state-of-the-art object detectors~\cite{he2017mask} failed on SDD. In this study, a motion detector~\cite{methylDragon} is adopted for SDD but does not perform well in challenging situations such as shaky camera and crowded area.
For pedestrian tracking, we utilize a popular tracker SORT~\cite{BewleyGORU16}
with the ``tracking-by-detection'' framework based on Kalman filter and conduct the proposed re-tracking algorithm upon SORT.
For trajectory prediction, we
use the ground truth trajectories to 
train an LSTM encoder-decoder model~\cite{AlahiGRRLS16} by using $L2$ loss and Adam optimizer with learning rate $1\times10^{-4}$ for 50 epochs (reduced by a factor of 5 at the 40th epoch).
The observation length for prediction is 4 time steps (1.6 sec).
The smoothing parameters in Eq.~\eqref{eq:holt} are set to $\alpha=\beta=0.5$.

To resolve the differences in scale among different videos, we convert the image coordinates (in pixel) of the center of each tracked bounding box to the world coordinates (in meter) with the given homography matrices and report the tracking and prediction performance in world coordinates.

\begin{figure*}[t]
\centering
\begin{tabular}{cc}
\hspace{-3mm}\includegraphics[height=2.61in]{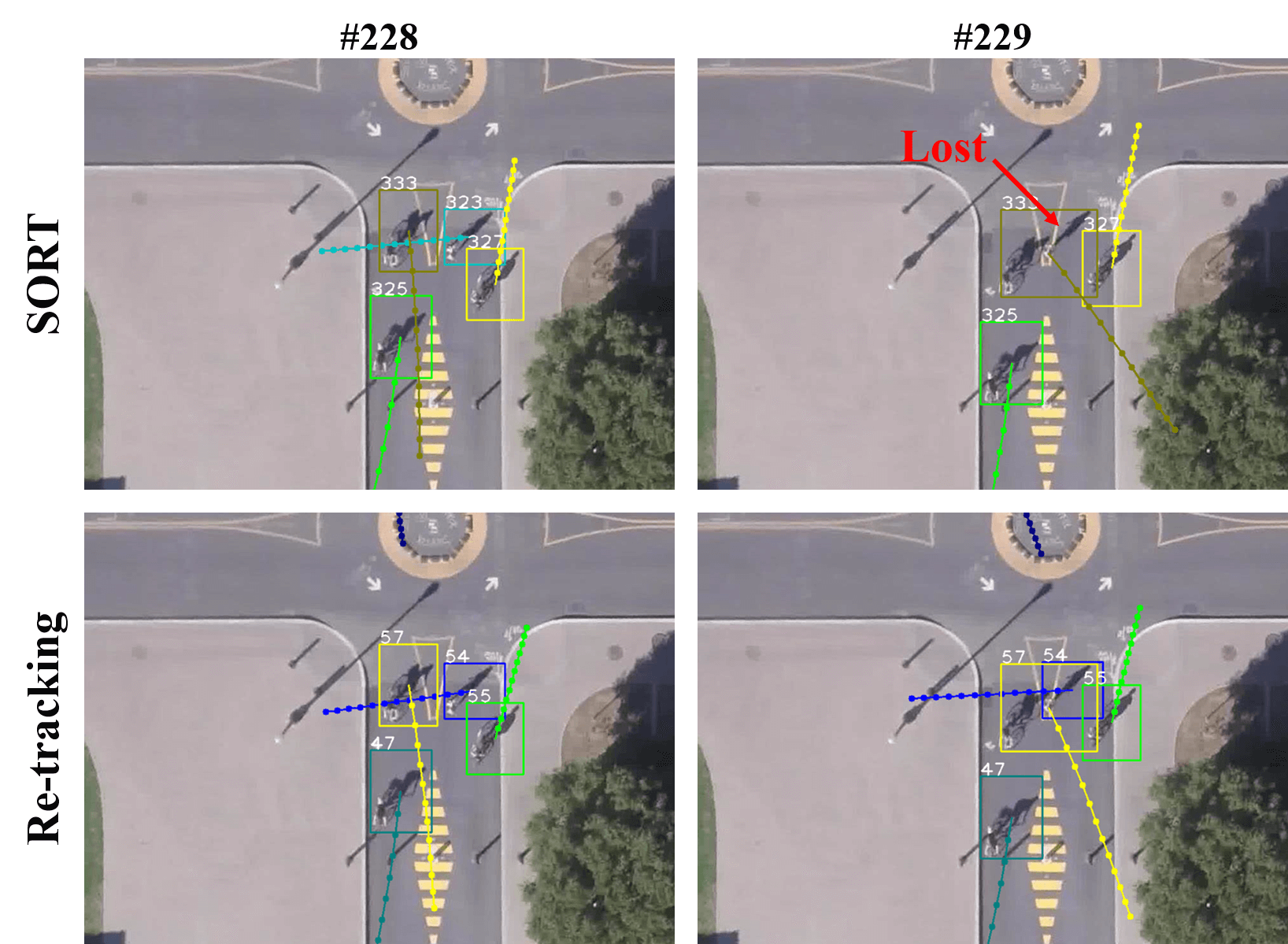} &
\hspace{-3mm}\includegraphics[height=2.61in]{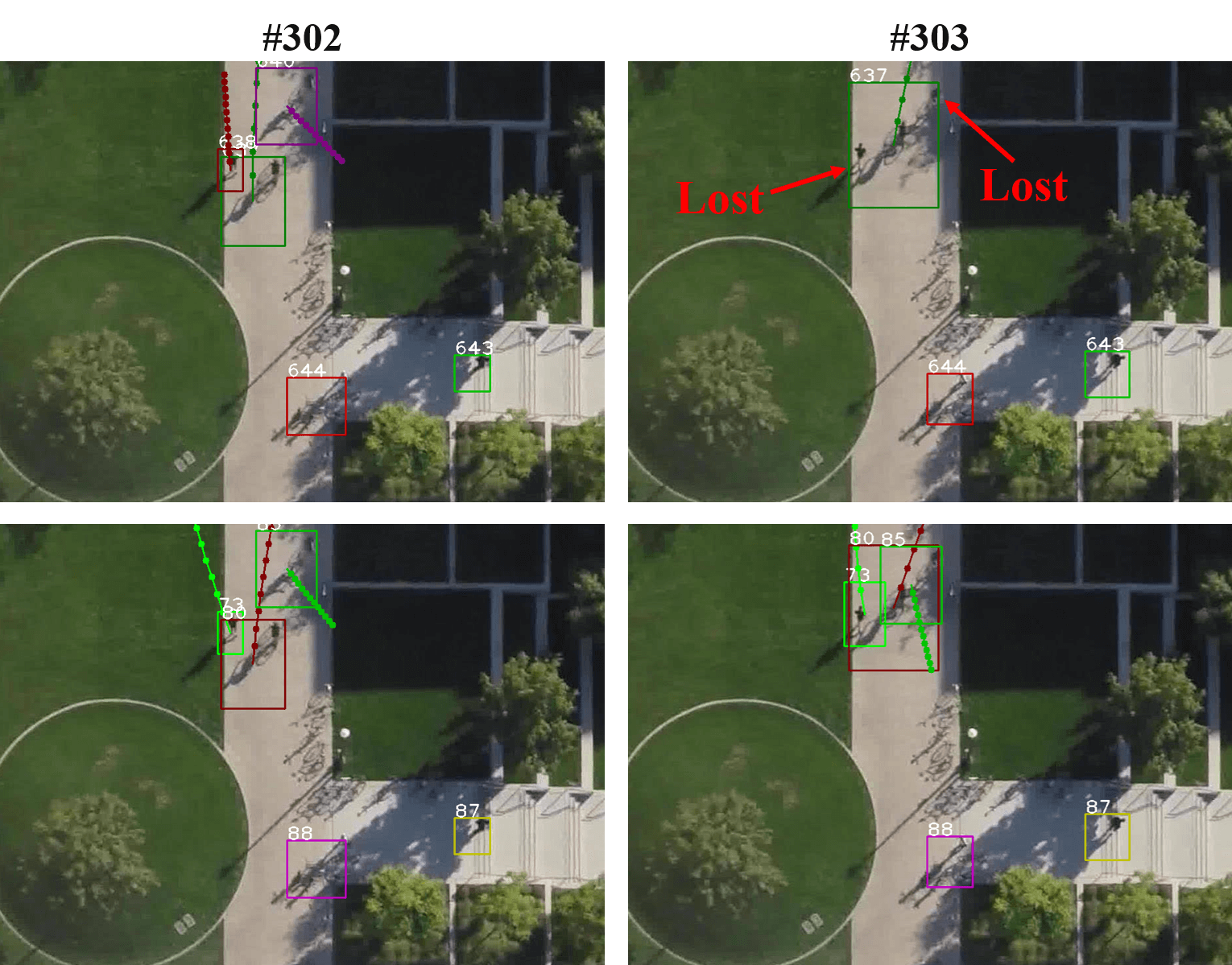} \\
{\small(a) Scene Little-3} & {\small(b) Scene Hyang-0}
\end{tabular}
\begin{tabular}{cc}
\hspace{-3mm}\includegraphics[height=2.61in]{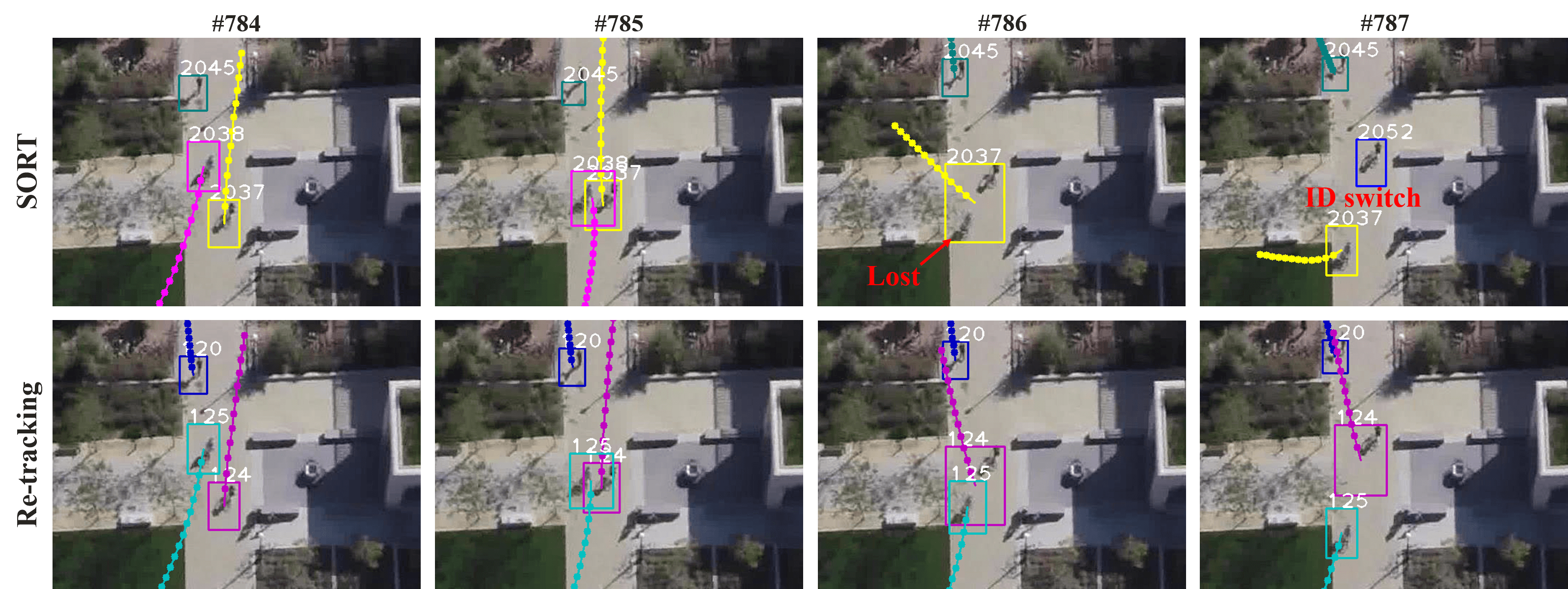} \\
{\small(c) Scene Hyang-3}
\end{tabular}
\begin{tabular}{cc}
\hspace{-3mm}\includegraphics[height=2.61in]{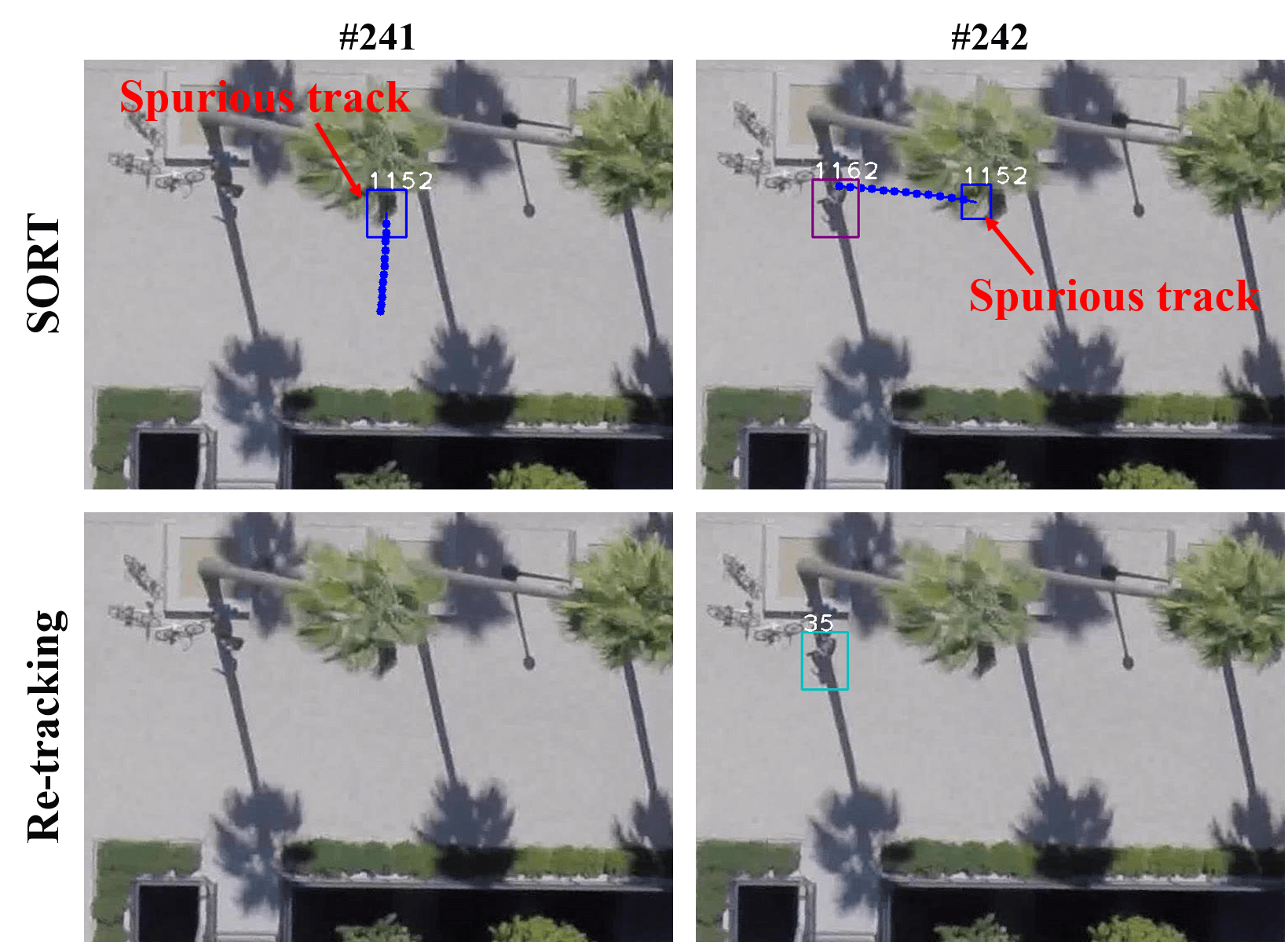} &
\hspace{-3mm}\includegraphics[height=2.61in]{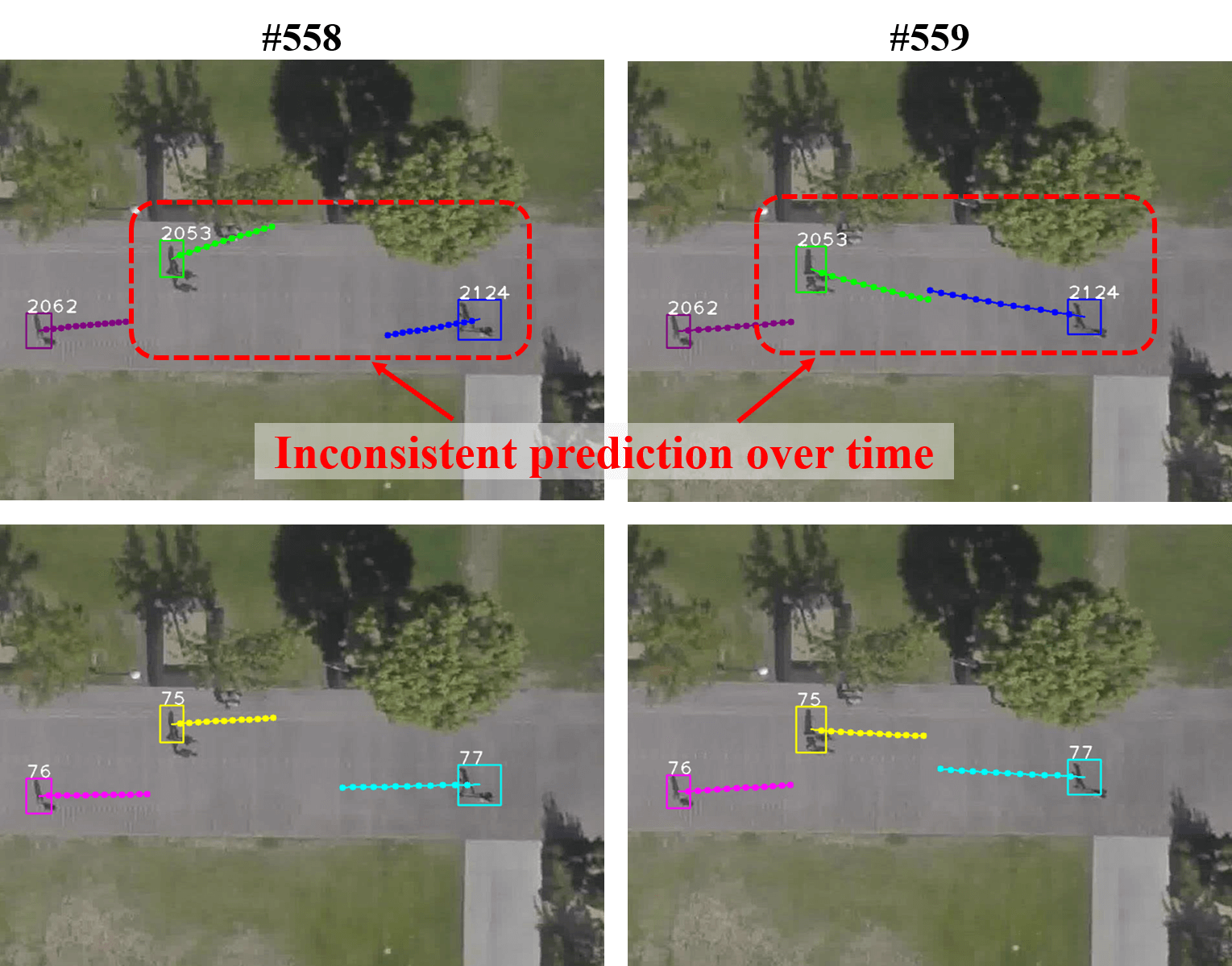} \\
{\small(d) Scene Coupa-0} & {\small(e) Scene Nexus-6}
\end{tabular}
\caption{Visualization of tracking and prediction at consecutive prediction time steps on SDD dataset. In each subfigure, the first row shows SORT~\cite{BewleyGORU16} results (baseline); the second row shows the re-tracking results. (a) SORT lost ID323 when the subject walked close to ID333. (b) SORT only maintained ID637 and lost the other two nearby subjects. (c) SORT lost ID2038 and switched to ID2037. (d) SORT generated a spurious track (ID1152) for a swaying tree. (e) Due to camera shake, the predictions based on SORT tracks were inconsistent over time, causing large prediction errors. The re-tracking method was able to handle the aforementioned problems.}
\label{fig:qualitative}
\end{figure*}

\subsection{Case Studies}


We first visualize the results on different SDD sequences to analyze the effect of re-tracking.
In each sub-figure of Fig.~\ref{fig:qualitative}, the first row shows the tracking results of SORT at consecutive prediction time steps; the second row shows those of re-tracking. The bounding boxes with ID numbers represent the tracks. Each prediction is denoted by a path with 12 future points. There are some boxes without predictions, due to insufficient number of past observations ($<4$). By default, our analysis will use the ID numbers of SORT. 

\smallskip
\noindent{\bf Missed targets.}
Fig.~\ref{fig:qualitative}(a) and (b) show examples where the re-tracking method correctly handles the missed targets.
In Fig.~\ref{fig:qualitative}(a), one can see that ID323 was crossing the road at Time \#228. 
At Time \#229, ID323 got close to ID333, and the detector only produced one large bounding box for the two persons. SORT allocated the box to ID333, thus lost the track of ID323. In contrast, as seen in the second row, our re-tracking method maintained both IDs at Time \#229. This is because the re-tracking method uses long-term predictions for association, thus was able to match track ID323 with a future track (of the same person) as the two people separate.

Fig.~\ref{fig:qualitative}(b) shows a similar situation of three persons. Since there was only one detection box at Time \#303, SORT only kept ID637 and lost the other two nearby subjects. The re-tracking methods maintained all tracks using the prediction-based association.

\smallskip
\noindent{\bf ID switches.}
Fig.~\ref{fig:qualitative}(c) shows an example where the re-tracking method avoids ID switches in the ``meet and separate'' situation. At Time \#784 and \#785, ID2037 and ID2038 approached each other. Then at Time \#786, the detector only generated one box, and SORT allocated it to ID2037 but lost ID2038. When the pedestrians separated at Time \#787, the detector produced two boxes for them again, but SORT associated ID2037 with the wrong one. As a result, the prediction of ID2037 at \#787 was very different from those at previous time steps. In contrast, the re-tracking method successfully recovered both tracks, because it preferred similar predictions (\ie, the prediction of ID2037 at \#785 and the prediction of ID2052 at \#787) during association.

\smallskip
\noindent{\bf Spurious tracks.}
The re-tracking method can also eliminate spurious tracks. As seen in Fig.~\ref{fig:qualitative}(d), the movement of a swaying tree introduced false detections, which were tracked by SORT as ID1152. However, the spurious track led to diverse predictions at Time \#241 and \#242. In our re-tracking algorithm, the two predictions could not be associated. As a result, the track ID1152 would be divided into short segments and subsequently deleted due to not surviving a probationary period. Note that SORT also utilizes a probationary period. But it relies on the bounding box distances for association, thus was unable to filter such false detections.

\smallskip
\noindent{\bf Noisy tracks.}
Finally, Fig.~\ref{fig:qualitative}(e) shows the effect of re-tracking on resolving the inconsistent predictions due to noisy tracks. 
In the scenes with even a slight camera shake (\eg, Nexus-6), the predictions from SORT tracks can be quite different at two consecutive time steps. As seen in Fig.~\ref{fig:qualitative}(e), the predictions of ID2053 and ID2124 at Time \#559 are very different from those at \#558. By smoothing the history trajectories, the same prediction model can generate more stable predictions from the re-tracking outputs, resulting in smaller prediction errors.

\begin{table}[t]
    \centering
    \caption{Tracking performance on SDD dataset.}
    \vspace{-2ex}
    \label{tab:MOT-result}
    \begin{tabular}{l c c c c c c c}
    \toprule
    Method & IDF1$\uparrow$ & MOTA $\uparrow$ & IDSW$\downarrow$ & FP$\downarrow$ & FN$\downarrow$\\
    \midrule
    SORT & 36.0 & 22.6 & 3,611 & 17,517 & 45,159\\
    Re-tracking & 44.9 & 30.1 & 1,246 & 11,441 & 47,242\\
    \bottomrule
    \end{tabular}
\end{table}

\subsection{Quantitative Results on SDD}

\noindent{\bf Tracking results.} As discussed before, tracking in SDD videos is difficult because of the small objects, crowd scenes, camera movements, and other factors. 
As seen in Table~\ref{tab:MOT-result}, the baseline method (SORT) yields high IDSW, FP, and FN numbers.
For example, shaky cameras bring a lot of false positives, whereas crowded areas lead to false negatives. Compared with SORT, our re-tracking approach increases the overall MOTA by 7.5 points (22.6 to 30.1) and IDF1 by 8.9 points (36.0 to 44.9). Most notably, it reduces IDSW by more than $65\%$ (3,611 to 1,246). The significant improvements in IDF1 and IDSW indicate that our re-tracking method yields more accurate associations, which in turn suggests that the proposed prediction-based distance metric Eq.~\eqref{eq:traj_dist} is more robust in real-world crowded scenes. Additionally, the decrease in FP (17,517 to 11,441) indicates that the re-tracking algorithm can effectively remove spurious tracks.


\setlength{\floatsep}{3pt}
\begin{figure}[t]
\centering
\includegraphics[width=0.65\linewidth]{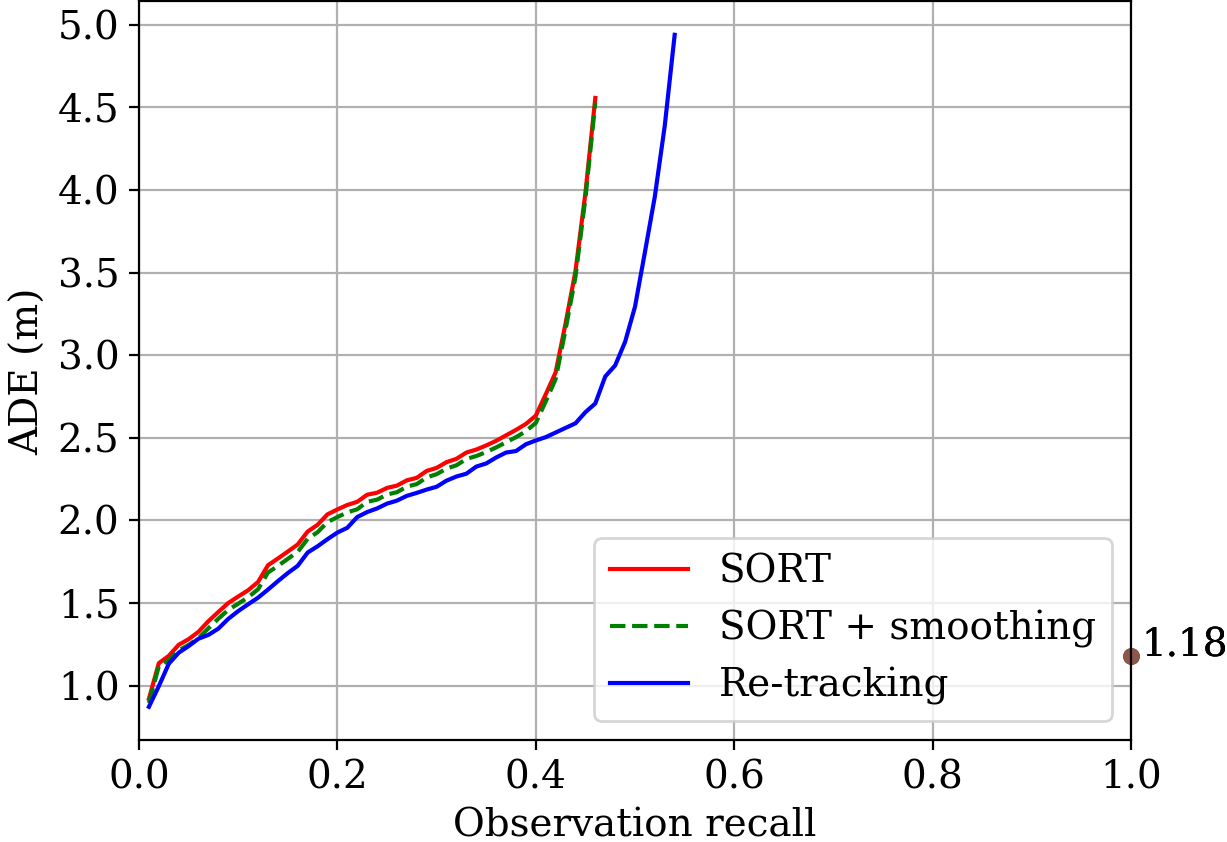}
\vspace{-2ex}
\caption{Comparison of ADE-over-recall curves on SDD testing sets.}
\label{fig:ade-curve-sdd}
\end{figure}


\smallskip
\noindent{\bf Prediction results.} Fig.~\ref{fig:ade-curve-sdd} shows the ADE-over-recall curves on the SDD testing sets, generated by using different threshold $\tau$ to associate the tracked subjects with the ground truth as in Eq.~\eqref{eq:association}. In agreement with the MOT metrics, our re-tracking method can achieve higher recall (53.7\%) than SORT (46.0\%). Besides, with the same LSTM model, the re-tracking method also yields smaller prediction ADE than SORT under the same observation recall values. Note that the improvements in prediction performance are derived mainly from two aspects: reduced ID switches and smoothed observations. As illustrated in Fig.~\ref{fig:qualitative}(c), ID switch can induce incorrect predictions, and the prediction model often makes inconsistent predictions over time with noisy tracks, as shown in Fig.~\ref{fig:qualitative}(e).

As a comparison, we also report the overall ADE for the prediction model given the ground truth observations. As shown in Fig.~\ref{fig:ade-curve-sdd}, prediction based on ``perfect'' tracking (\ie, 100\% recall) yields an ADE of 1.18. However, when taking the tracking results as input, the prediction model can achieve a comparable ADE only at very low recall levels (\ie, $<10\%$). This clearly shows that there still exists a large room for future improvements.

\begin{figure}[t]
\centering
\includegraphics[width=1.0\linewidth]{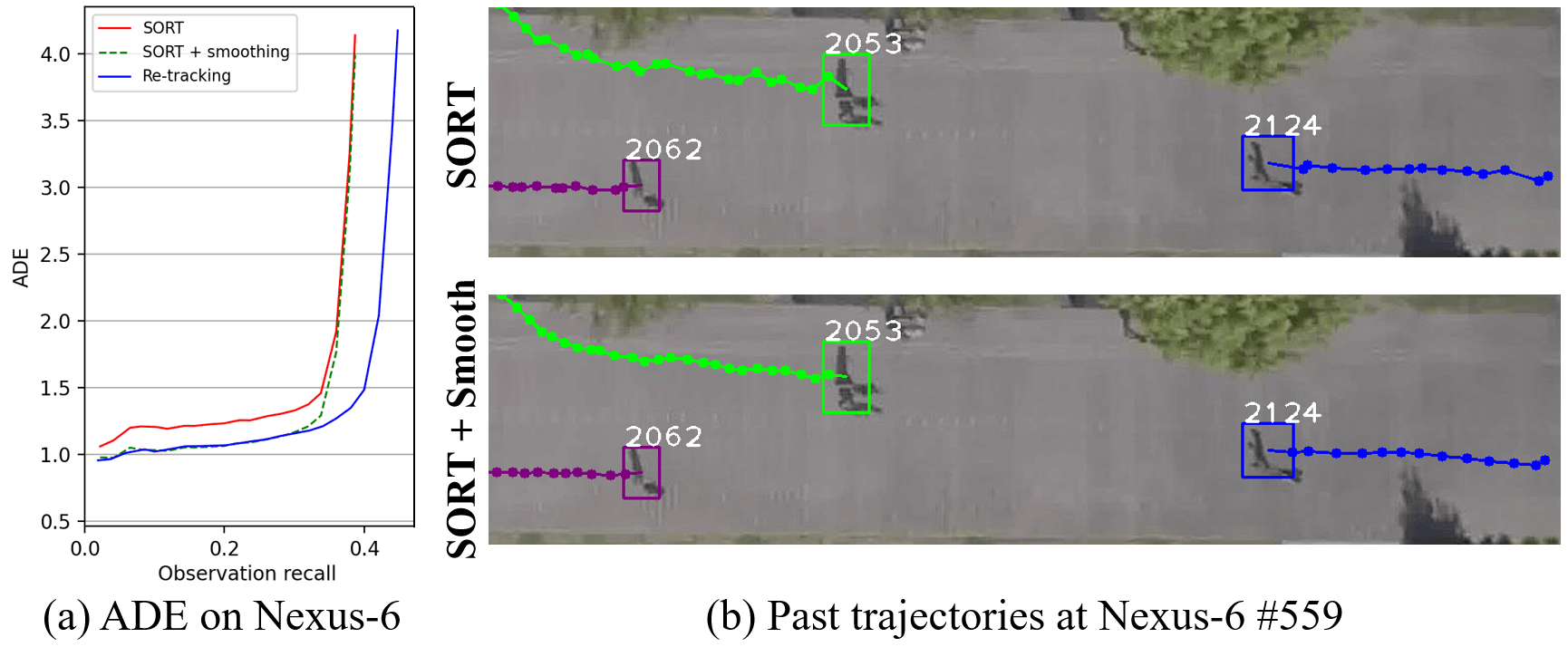}
\vspace{-5ex}
\caption{Analysis of smoothing effect. (a) Comparison of ADE-over-recall curves on Nexus-6. (b) Visualization of past trajectories (1st row: original SORT; 2nd row: smoothed SORT.)}
\label{fig:smooth-effect}
\end{figure}

\smallskip
\noindent{\bf Effect of smoothing.}
As an ablation study, we also evaluate the effect of track smoothing on trajectory prediction. We directly apply Eq.~\eqref{eq:holt} to every SORT track and re-evaluate the prediction performance. The overall ADE-over-recall curve is shown in Fig.~\ref{fig:ade-curve-sdd} with the label ``SORT+smoothing''. By smoothing the tracks, the ADE drops slightly at different recall levels. The mean decrease in ADE across all recalls is 0.040 with a standard deviation of 0.006.

For individual videos with camera shake (\eg, Nexus-6), smoothing the tracks can significantly improve the prediction performance, as shown in Fig.~\ref{fig:smooth-effect}(a). The mean decrease in ADE across all recalls is 0.158 with a standard deviation of 0.020. We visualize the history SORT tracks at prediction time \#559 on Nexus-6 in the first row of Fig.~\ref{fig:smooth-effect}(b). As analyzed in~\cite{SchollerALK20}, the prediction of neural networks heavily relies on the most recent two points, which explains the correlation between the SORT tracks in Fig.~\ref{fig:smooth-effect}(b) and the inconsistent predictions in Fig.~\ref{fig:qualitative}(c). The proposed smoothing method can reduce the observation noise level, thus improves the prediction consistency.

\smallskip
\noindent{\bf Runtime.}
We conducted experiments on a desktop with an Intel i7-7700 CPU, 32GB RAM, and an Nvidia Titan XP GPU. The entire pipeline (including SORT, re-tracking, and prediction) takes about 0.004s to process a frame, thus is suitable for real-time, online applications.



\subsection{Discussion}
\label{sec:discussion}

We have demonstrated that, by considering prediction consistency during re-tracking, it is possible to achieve better tracking and prediction results. However, as shown in Fig.~\ref{fig:ade-curve-sdd}, the ADE increases rapidly as the recall increases, especially when recall $>0.4$. 
A close look at the experiment results reveals that recall $=0.4$ approximately corresponds to setting the threshold $\tau=2.0$ for association. For the SDD dataset, if the average distance between two instances is larger than $2.0$, it is very unlikely that the two instances belong to the same subject. In other words, the association tends to be wrong, which explains the large errors in the long-term prediction for recall $>0.4$. Meanwhile, for threshold $\tau \in [0, 1.5]$ (\ie, mostly correct associations), the ADE roughly falls in the range $[1.0,2.5]$, which is approximately equal to the observation error $[0,1.5]$ plus 1.18 (\ie, the ADE obtained using ground truth trajectories).

Therefore, to further improve the ADE-over-recall curve, the key is to increase the number of correctly tracked subjects. Although our re-tracking method increases the recall from 46.0\% to 53.7\%, it is still limited by the original detection and tracking algorithms. In particular, the re-tracking method will not recover subjects that are not tracked by SORT for a long period of time. This is evident by the large number of false negatives (FN) in Table~\ref{tab:MOT-result} for both SORT and our re-tracking method. Besides, compared to SORT, using re-tracking has two opposite effects on FN: (1) It can reduce FN by recovering missed targets, as shown in Fig.~\ref{fig:qualitative}(a)-(c). (2) Since re-tracking method relies on future predictions, it will discard short tracks where predictions are unavailable, thus possibly increase FN.




To further improve the recall, one possible direction is to learn a joint model for detection, tracking, and prediction. The intuition is that, future predictions can be used to infer the subject's location during tracking, whereas accurate tracking results can improve the prediction accuracy. Recently, several work proposed to perform such joint inference on point cloud data~\cite{LuoYU18,CasasLU18,GLU20,LiangYZCH0U20,CasasGSLLU20,LiYLZRSU20,abs-2010-00731}. However, one challenge in applying similar ideas to datasets like SDD is that state-of-the-art data-driven detectors~\cite{he2017mask} failed to detect small objects in bird's-eye view. In this work, we resort to a motion-based detector instead. It is interesting to develop a model that leverages the motion cues in videos for joint detection, tracking, and prediction.


Alternatively, one may learn a prediction model that is more robust to tracking errors and noises. For example, since noisy tracks often lead to inconsistent predictions, one could compute the difference in predictions at consecutive time steps and use it as a loss term for training the model. 




\subsection{Results on WILDTRACK}
\label{sec:wildtrack}
As further verification, we also conduct experiments on WILDTRACK~\cite{ChavdarovaBBMJB18}, a large-scale dataset for multi-camera pedestrian detection, tracking, and trajectory forecasting~\cite{abs-2007-03639}. It consists of 7 videos ($\sim$35 min) captured by 7 calibrated and synchronized static cameras (60 fps) in a crowded open area in ETH Zurich.
The original dataset annotated 7$\times$400 frames at 2 fps. We annotated additional 7$\times$500 frames, resulting in 94,361 annotations in total.
In this study, 7$\times$600 frames are used for training and the rest 7$\times$300 for testing.
Different from SDD, the WILDTRACK cameras were mounted at the average human height, and we are able to obtain satisfactory pedestrian detection results using a pretrained Mask R-CNN~\cite{he2017mask} model. We conduct SORT tracking at 10 fps and make predictions with LSTM model at 2 fps for future 9 time steps (4.5 sec) based on 3-step observations (1.5 sec). The main challenge in WILDTRACK is the occlusions among pedestrians in the crowd.

\begin{table}[t]
    \centering
    \caption{Tracking performance on WILDTRACK dataset.}
    \vspace{-2ex}
    \label{tab:MOT-result-wildtrack}
    \begin{tabular}{l c c c c c c c}
    \toprule
    Method & IDF1$\uparrow$ & MOTA $\uparrow$ & IDSW$\downarrow$ & FP$\downarrow$ & FN$\downarrow$\\
    \midrule
    SORT & 41.4 & 14.1 & 1,182 & 12,117 & 14,454\\
    Re-tracking & 43.4 & 15.0 & 654 & 10,713 & 16,083\\
    \bottomrule
    \end{tabular}
\end{table}

\begin{figure}[t]
\centering
\includegraphics[width=0.65\linewidth]{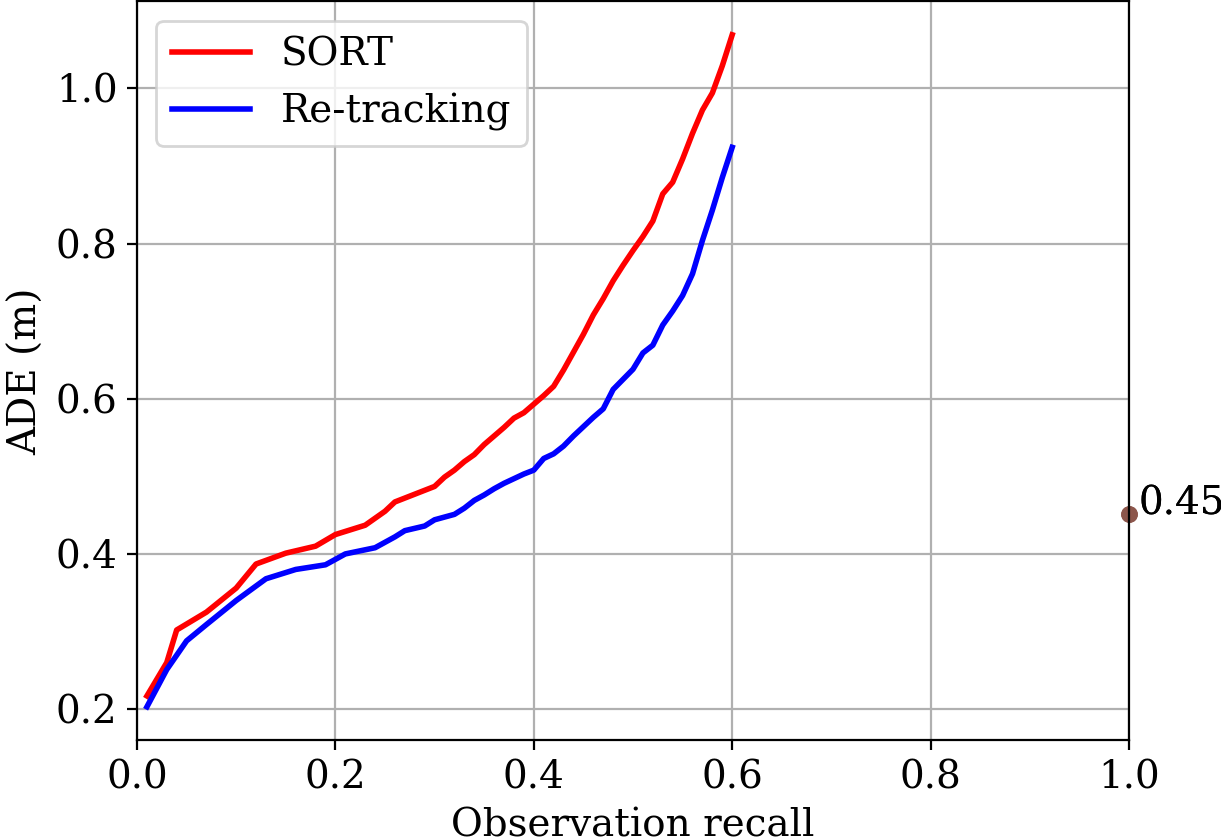}
\vspace{-2ex}
\caption{Comparison of ADE-over-recall curves on WILDTRACK}
\label{fig:ade-curve-wildtrack}
\end{figure}

We report quantitative results on WILDTRACK in Table~\ref{tab:MOT-result-wildtrack} and Fig.~\ref{fig:ade-curve-wildtrack}, and refer readers to supplementary video for visualizations. As seen in Table~\ref{tab:MOT-result-wildtrack}, re-tracking improves the tracking performance in IDF1 (41.4 to 43.4), MOTA (14.1 to 15.0), IDSW (1,182 to 654), and FP (12,117 to 10,713). The significant drop in IDSW (about 45\%) again reflects the strength of our method in creating more accurate associations. According to Section~\ref{sec:discussion}, we only report the ADE-over-recall curve where the threshold $\tau\leq2.0$ (corresponding to recall $\leq 0.6$) in Fig.~\ref{fig:ade-curve-wildtrack}. At every recall level, the LSTM model achieves a lower ADE based on the re-tracking trajectories.
The improvement in both tracking and prediction performance on WILDTRACK shows that the re-tracking method is not only effective in bird's-eye videos (\eg, SDD) but also eye-level videos (\eg, WILDTRACK).

\section{Conclusion}
In this paper, we study human trajectory forecasting in raw videos. We carefully analyze how false tracks and noisy trajectories affect prediction accuracy. We illustrate the importance of temporal consistency in prediction, and propose a ``re-tracking’’ algorithm to enforce prediction consistency over time. Through case studies, we demonstrate that our re-tracking algorithm can address different types of tracking failures. On the SDD and WILDTRACK benchmark datasets, the proposed method consistently boosts both the tracking and prediction performance.
As one of the first attempts to bridge the gap between tracking and prediction, this study leaves several research opportunities for human trajectory prediction in raw videos.

\bibliographystyle{IEEEtran}
\bibliography{iros21}
\end{document}